\documentclass[letterpaper, 10 pt, conference]{ieeeconf}  %

\IEEEoverridecommandlockouts                              %

\usepackage{graphicx} %
\usepackage{times} %
\usepackage{amsmath} %
\usepackage{amssymb}  %
\usepackage[colorlinks]{hyperref}
\usepackage{color}
\usepackage[dvipsnames]{xcolor}
\usepackage{caption}
\usepackage{float}
\usepackage{subcaption}
\usepackage{titlesec}
\titlespacing{\section}{0pt}{2mm}{1mm}
\titlespacing{\subsection}{0pt}{1mm}{0.25mm}
\titlespacing{\subsubsection}{0pt}{0mm}{0mm}

\newtheorem{problem}{Problem}

\title{\LARGE \bf
 Legible and Proactive Robot Planning for\\Prosocial Human-Robot Interactions
}

\author{Jasper Geldenbott $\quad$ Karen Leung%
\thanks{This work was supported by the UW + Amazon Science Hub Research Award.}%
\thanks{University of Washington, Department of Aeronautics and Astronautics
        {\tt\small \{jgelden, kymleung\}@uw.edu }}%
\thanks{We would like to acknowledge Purnanand Elango, Annika Singh, Chris Hayner, and Nia Jetter for their insightful discussions on trajectory optimization, and Isaac Remy for his help on the experiments.}
}

\newcommand{\uh}{\mathbf{u}_\mathrm{H}}         %
\newcommand{\xht}[1]{\mathbf{x}_\mathrm{H}^{#1}}         %
\newcommand{\uht}[1]{\mathbf{u}_\mathrm{H}^{#1}}         %
\newcommand{\xrt}[1]{\mathbf{x}_\mathrm{R}^{#1}}         %
\newcommand{\urt}[1]{\mathbf{u}_\mathrm{R}^{#1}}         %

\newcommand{\xt}[1]{\mathbf{x}^{#1}}         %
\newcommand{\ut}[1]{\mathbf{u}^{#1}}         %
\newcommand{\traj}[1]{\mathbf{\tau}^{#1}}         %

\newcommand{\xjt}[1]{\mathbf{x}_\mathrm{HR}^{#1}}         %
\newcommand{\fj}{f_\mathrm{HR}}         %

\newcommand{\xlt}[1]{\mathbf{x}_\mathrm{L}^{#1}}         %
\newcommand{\ult}[1]{\mathbf{u}_\mathrm{L}^{#1}}         %
\newcommand{\xf}{\mathbf{x}_\mathrm{F}}         %
\newcommand{\uf}{\mathbf{u}_\mathrm{F}}         %
\newcommand{\xft}[1]{\mathbf{x}_\mathrm{F}^{#1}}         %
\newcommand{\uft}[1]{\mathbf{u}_\mathrm{F}^{#1}}         %
\newcommand{\ff}{f_\mathrm{F}}         %

\begin{document}

\maketitle
\thispagestyle{empty}
\pagestyle{empty}

\begin{abstract}
    Humans have a remarkable ability to fluently engage in joint collision avoidance in crowded navigation tasks despite the complexities and uncertainties inherent in human behavior.
    Underlying these interactions is a mutual understanding that (i) individuals are \textit{prosocial}, that is, there is equitable responsibility in avoiding collisions, and (ii) individuals should behave \textit{legibly}, that is, move in a way that clearly conveys their intent to reduce ambiguity in how they intend to avoid others.
    Toward building robots that can safely and seamlessly interact with humans, we propose a general robot trajectory planning framework for synthesizing legible and proactive behaviors and demonstrate that our robot planner naturally leads to prosocial interactions. Specifically, we introduce the notion of a \textit{markup factor} to incentivize legible and proactive behaviors and an \textit{inconvenience budget constraint} to ensure equitable collision avoidance responsibility.
    We evaluate our approach against well-established multi-agent planning algorithms and show that using our approach produces safe, fluent, and prosocial interactions. We demonstrate the real-time feasibility of our approach with human-in-the-loop simulations. Project page can be found at \url{https://uw-ctrl.github.io/phri/}.
    \end{abstract}

    \section{Introduction}
    Robots are becoming increasingly pervasive in our everyday lives, and more so in settings where they must interact and navigate through human counterparts. 
    From autonomous driving to autonomous food deliveries, robots are required to safely and seamlessly interact with humans despite the myriad of uncertainties they face such as human intent, human preferences, and environmental factors. Indeed, a significant research effort has been dedicated to human behavior prediction (see \cite{RudenkoPalmieriEtAl2020} for a survey).
    Yet in highly dense interactions (e.g., Shibuya Crossing in Tokyo, Japan), humans are surprisingly remarkable in avoiding collision. 
    The reason is, that humans are self-preserving \cite{MayerBellEtAl2021} and will engage in \textit{joint collision avoidance}. Joint collision avoidance controllers have been shown to successfully replicate collision-free and fluent (dense) multi-agent interactions \cite{HelbingMolnar1995,VandenBergLinEtAl2008,KnepperRus2012}.
    While we may not be able to perfectly predict how humans behave, we can at least expect other humans to exhibit some degree of cooperativeness in avoiding collisions.
    In this work, we ask the question, \textit{despite uncertainty in human behaviors, can we achieve safe and fluent human-robot interactions by leveraging the fact that humans are self-preserving?} 
    
    Our main hypothesis is that if the robot is able to indicate \textit{early} to interacting humans its \textit{intent} to avoid collision (e.g., pass to the right), then this will (i) provide the humans sufficient warning to adjust their plans to avoid collisions,  (ii) remove ambiguity in how the joint collision avoidance maneuver should occur, and (iii) result in \textit{prosocial} interactions where everyone equitably compromises their performance to benefit the group. 
    In this work, we propose a robot planning framework to create \textit{legible}---the ability to infer the intent from motion---and \textit{proactive}---the ability to influence a situation by causing something to happen rather than responding to it after it has happened---robot motions.
    Our approach adds a number of simple additions to a standard robot trajectory optimization used widely for robot motion planning, making it a general and flexible approach for synthesizing legible and proactive behaviors. Specifically, we introduce the notion of a \textit{markup} factor to encourage the robot to take nontrivial actions earlier rather than later, and the concept of an \textit{inconvenience budget} to keep the robot's motion directed to its goal and to promote an equitable distribution of collision avoidance responsibility.

    \begin{figure}
        \centering
        \includegraphics[width=0.48\textwidth]{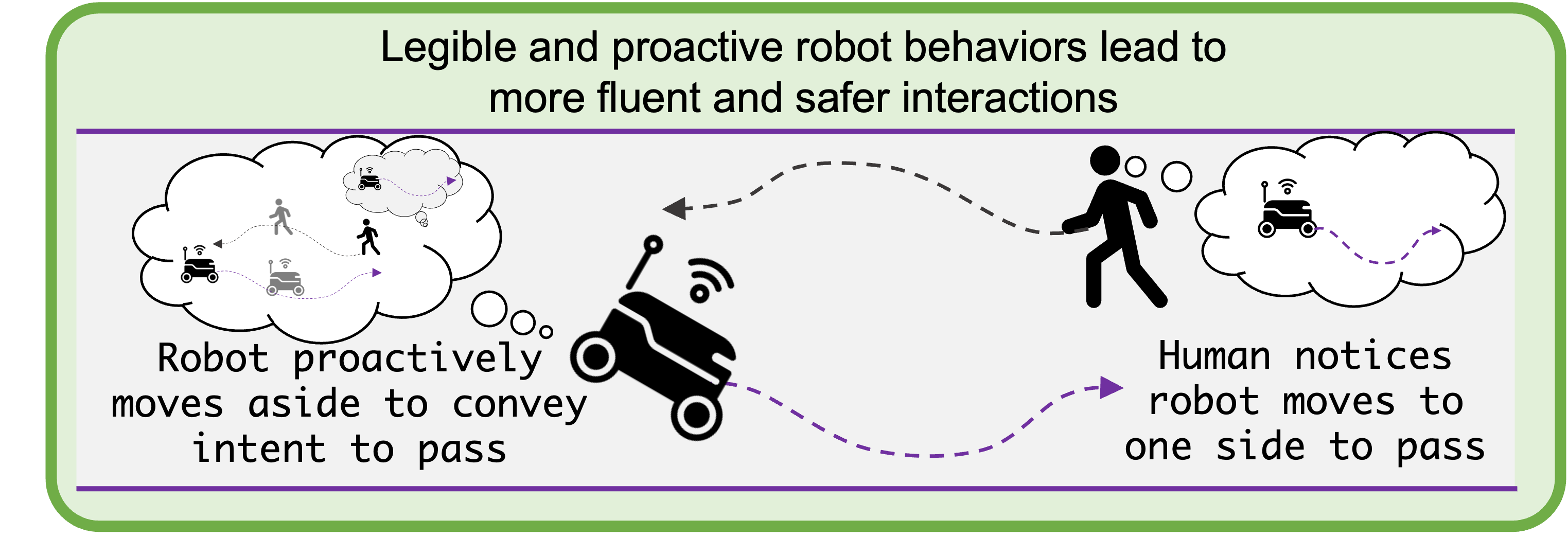}
        \caption{Our approach enables robots to interact legibly and proactively, providing oncoming humans with sufficient time to engage cooperatively in collision avoidance. Our experiments show that this leads to fluent, safe, and prosocial interactions. 
        }
        \label{fig:hero}
        \vspace{-7mm}
    \end{figure}

    \noindent \textbf{Contributions.} Our contributions are four-fold: 
    (i) We introduce an inconvenience budget constraint as a mechanism to produce prosocial interactions. The constraint prevents the robot from making inequitable collision avoidance maneuvers, such as a sharp swerve, and keeps its motion aligned with the goal direction.
    (ii) We present a markup factor into the robot's cost function to encourage legible and proactive behaviors. With the markup factor, the robot is encouraged to rotate earlier to pass oncoming agents, therefore indicating its intent to the human.
    (iii) We propose a robot planner utilizing an iterated best response algorithm to capture interaction dynamics with human agents and demonstrate its real-time applicability via human-in-the-loop experiments.
    (iv) We demonstrate that our proposed legible and proactive planner results in safer, more fluent, and more prosocial interactions compared to other social navigation approaches.
    
    \noindent \textbf{Organization.} 
    In Section~\ref{sec:related work}, we discuss related work, followed by a problem formulation in Section~\ref{sec:problem formulation}. We describe our proposed approach in Section~\ref{sec:methodology}, and discuss experimental results in Section~\ref{sec:experiment and discussion}. We conclude and present exciting directions for future work in Section~\ref{sec:conclusion}.
    
    \section{Related work}
    \label{sec:related work}
    We discuss some related work on social robot navigation, with a focus on modeling other agents, social competency, and legibility. For a deeper review, see \cite{MavrogiannisBaldiniEtAl2023} and \cite{FrancisPerezD’ArpinoEtAl2023}.

    \noindent \textbf{Modeling other agents.} To model feedback interaction between robots and humans, the robot typically relies on knowledge of humans' objective (cost or reward) function, often chosen to be a linear combination of handcrafted features \cite{SadighSastryEtAl2016c,KretzschmarSpiesEtAl2016,SunZhanEtAl2018}.
    A key benefit is interpretability and tractability---knowledge of the objective function provides direct insight into what other agents value and how they would respond in any given situation (assuming they are optimal planners).
    However, data is required to learn other humans' objective function; employing techniques such as maximum entropy inverse reinforcement learning \cite{ZiebartMaasEtAl2008,LevineKoltun2012}. 
    While these approaches boast interpretability benefits, the reliance on knowing other humans' objective functions hinders real-world multi-agent applicability. A deep neural network can be used to learn the objective function but would lose any interpretability benefits \cite{FinnLevineEtAl2016}. Online inference is challenging given the limited interaction time while offline learning is prone to distribution shifts. 
    In our work, we do not rely on learning other humans' objective functions from data. Instead, we propose a robot-only objective function that encourages \textit{legible} and \textit{proactive} behaviors to help the robot indicate to humans early its intent. We show that reducing ambiguity in robot behaviors helps humans commit to a type of behavior \cite{CartonOlszowyEtAl2016}, thereby mitigating the reliance on accurate feedback interaction models.

    \noindent \textbf{Social competency.} To increase interpretability and structure into understanding how robots can account for the welfare of others, the concept of Social Value Orientation (SVO) stems from social psychology which models the degree to which an individual values other people’s welfare in relation to their own \cite{LiebrandMcClintock1988}. Different SVO parameter values elicit different behaviors, such as egoistic, prosocial, altruistic, masochistic, and sadistic.
    The SVO parameter can be learned from data to provide a social label on interaction data \cite{SchwartingPiersonEtAl2019,CrosatoShumEtAl2023}, or chosen by the designer to elicit desired robot behaviors \cite{ToghiValienteEtAl2021,ToghiValienteEtAl2022}. 
    Building upon the idea of socially aware robots, \cite{SunZhanEtAl2018,SchaeferLeungEtAl2021} consider designing ``courteous'' robot behaviors by minimizing the amount of inconvenience the robot imposes on the human. Specifically, they compare how humans behave when the robot is there versus if the robot were not there. Similar to the works discussed previously, \cite{SunZhanEtAl2018,SchaeferLeungEtAl2021} too operate in the objective space---a major limitation to encoding social preference in the objective space is that it does not limit how much a robot should compromise its objective for the benefit of others. Without such a limit, it may lead to the robot behaving in unexpected ways, such as moving backward \cite{SadighSastryEtAl2016c}, or becoming incapable of accomplishing the task to yield for others, i.e., frozen robot problem \cite{TrautmanKrause2010}.
    In our work, we impose a robot-centric \textit{inconvenience budget constraint}, and in conjunction with legible and proactive planning, our approach naturally leads to \textit{prosocial} behaviors compared to other baseline approaches.

    \noindent \textbf{Legibility.}
    There has been considerable attention into constructing legible robot behaviors and evaluating its influence in generating more fluent and safer human-robot interactions \cite{CartonOlszowyEtAl2016,DraganLeeEtAl2013,DraganSrinivasa2013,KruseBasiliEtAl2013,MavrogiannisAlves-OliveraEtAl2021,LichtenthaelerLorenzyEtAl2012,BuschGrizouEtAl2017}. 
    However, different approaches formulate legibility differently, leading to vastly different, highly customized legible motion planning algorithms that are difficult to integrate into a more general planning framework that considers additional planning objectives. For example, \cite{DraganSrinivasa2013} proposed a gradient-based legible motion optimizer for (single-agent) goal-reaching problems, and \cite{MavrogiannisAlves-OliveraEtAl2021} introduces the notion of social momentum to generate legible motion to indicate which side agents will pass by each other. 
    Leveraging the notion that the velocity vector is a primary signal for indicating intent \cite{SchoellerAravantinosEtAl2020}, we propose a simple approach to synthesizing legible robot behaviors that can be easily integrated into a general optimal control problem.

    \section{Problem formulation}
    \label{sec:problem formulation}
    We aim to compute robot controls that enable a robot to accomplish its task in a safe and efficient manner while taking into account the presence and reactions of surrounding human agents. For simplicity, first consider a single human case; we later describe how to extend this to multiple humans. Let $\xjt{t}\in\mathbb{R}^{n_\mathrm{H}+n_\mathrm{R}}$ denote the joint state of the human and robot $\xht{t}\in\mathbb{R}^{n_\mathrm{H}}$ and $\xrt{t}\in\mathbb{R}^{n_\mathrm{R}}$ at time step $t$ respectively, and $\uht{t}\in\mathbb{R}^{m_\mathrm{H}}$ and $\urt{t}\in\mathbb{R}^{m_\mathrm{R}}$ be the human and robot controls at time step $t$ respectively. Then we seek to find a robot control sequence $\urt{0:T}$ over $T$ time steps that is a solution to the following optimal control problem:
    
    \begin{problem}[Interaction-aware trajectory problem]
    \begin{subequations}
    \begin{align}
        \min_{\urt{0:T}}\; & \sum_{t=0}^T J(\xjt{t}, \urt{t}, t) + J_{T+1}(\xjt{T+1}) \label{eq:general:objective}\\
        \mathrm{s.t.} \; & \xjt{t+1} = \fj(\xjt{t}, \urt{t}, \uh(\xjt{t})),  t=0,...,T \label{eq:general:dynamics}\\
        & g_i(\xjt{t}, \urt{t}) \leq 0,  t=0,...,T, i=0,...,G \label{eq:general:inequality}\\
        & h_j(\xjt{t}, \urt{t})=0,  t=0,...,T, j=0,...,H \label{eq:general:equality}.
    \end{align}
    \label{eq:general formulation}
    \end{subequations}
    \label{prob:general interaction problem}
    \vspace{-7mm}
    \end{problem}
    \noindent The objective (cost) function \eqref{eq:general:objective} captures the robot's performance goals (e.g., minimize control effort and distance to the goal) and joint performance between both agents (e.g., the robot considers the welfare of the human). The joint dynamics constraints \eqref{eq:general:dynamics} capture the interaction between the two agents; the human's control policy $\uh(\xjt{t})$ is expressed as a function of the joint state---how the robot behaves will affect how the human behaves. Indeed, this joint dynamics can be viewed as an underactuated system---a large challenge is in obtaining realistic human policies and formulating a tractable robot trajectory planner with it. The other two constraints \eqref{eq:general:inequality} and \eqref{eq:general:equality} describe constraints on robot controls and the joint state such as initial state constraints, control constraints, and critically, collision avoidance.
    
    \section{Legible robot planning for proactive and prosocial behavior}
    \label{sec:methodology}
    
    We present our legible robot trajectory planner that encourages the robot to be legible and proactive, leading to prosocial human-robot interactions. The planner will be executing in model predictive control (MPC) fashion, recomputing a new solution at each time step.
    We leverage the notion that in the context of crowded social navigation, humans engage in \textit{joint collision avoidance} but are also self-interested in that they are unwilling to disproportionally sacrifice their own performance for the benefit of others.  
    Solving Problem~\ref{prob:general interaction problem} is challenging due to the closed-loop coupling between agents and difficulty in accessing the human's policy.
    Before diving into the robot planner, we first describe iterated best response (IBR) to account for the closed-loop interactions, a commonly used approach for interaction-aware planning \cite{SunZhanEtAl2018,SadighSastryEtAl2016c,WangWangEtAl2021}. 
    Although IBR requires a model for the human, we only assume knowledge of human dynamics and goal location.
    In general, our proposed robot planner is compatible with other interaction planners that do not require an explicit human model (e.g., neural networks) \cite{SchmerlingLeungEtAl2018,SchaeferLeungEtAl2021}, and we defer investigating these models for future work.
    
    \subsection{Iterated best response}
    We utilize an iterated best response (IBR) strategy to account for the human's reaction to the robot's actions.
    In IBR, the robot iteratively improves its trajectory by computing how the human would respond to its trajectory, and then vice versa.
    Without loss of generality, we first assume the human is the ``leader'' whose trajectory is fixed. Then the robot (the ``follower'') selects its trajectory after observing the leader's trajectory. Then the roles are switched and the agents update their trajectories correspondingly. This process repeats until convergence or a fixed number of iterations.
    The final output is the robot's trajectory, and the first control output will be executed as part of the MPC loop.
    The robot assumes that the human is a ``reasonable'' decision-maker who, like the robot, minimizes a cost function while avoiding collision.

    \subsection{Inner trajectory optimization}
    In this section, we describe the inner trajectory optimization problem during the IBR iterations. 
    Note that while both the human and robot solve the same trajectory optimization problem described in Problem~\ref{prob:follower trajectory problem}, the parameter values may differ for each agent.
    Without loss of generality, at any iteration of the IBR loop, the follower has observed the leader's planned trajectory $(\xlt{0:T+1}, \ult{0:T}$) which is considered fixed. The follower responds by selecting a state/control trajectory $(\xft{0:T+1},\uft{0:T})$ that is a solution to Problem~\ref{prob:follower trajectory problem},
    
    \begin{problem}[Follower's trajectory optimization problem]
    \allowdisplaybreaks
        \begin{subequations}
        {\small
    \begin{align}
    \min_{\substack{\xft{0:T+1},\\\uft{0:T},\\{\color{RoyalBlue}\epsilon_{0:T+1}}}} & \sum_{t=0}^T {\color{WildStrawberry}\mu^t}J(\uft{t}, \xft{t}, t) + \gamma_0\sum_{t=0}^{T+1}\gamma^t {\color{RoyalBlue}\epsilon_t}^2 + J_{T+1}(\xft{T+1}) \label{eq:follower:objective}\\
        \mathrm{s.t.} \quad & \xft{t+1} = \ff(\xft{t}, \uft{t}), \; \xft{0} = \xft{\mathrm{current}}\quad  t=0,...,T \label{eq:follower:dynamics}\\
        & \xft{t} \in \mathcal{X}_\mathrm{F}^t \setminus \mathcal{O}_\mathrm{static}, \quad t=0,...,T+1, \label{eq:follower:state constraint}\\
        & \uft{t} \in \mathcal{U}_\mathrm{F}(\xft{t}) , \quad t=0,...,T   \label{eq:follower:control constraints}\\
        & {\color{RoyalBlue}g(\xft{t},\uft{t}, \xlt{t},\ult{t}) \geq -\epsilon_t} , \quad t=0,...,T \label{eq:follower:collision avoidance}\\
        & {\color{RoyalBlue}g(\xft{T+1}, \xlt{T+1}) \geq -\epsilon_{T+1}} ,  \label{eq:follower:collision avoidance final}\\
        & {\color{OliveGreen}J_\mathrm{incon}(\xft{0:T+1},\uft{0:T}) \leq \beta_\mathrm{F}} , \label{eq:follower:inconvenience budget}\\
        & {\color{RoyalBlue}\epsilon_t \geq 0 }, \quad t=0,...,T+1.\label{eq:follower:slack}
    \end{align}
    \label{eq:follower formulation}
    }
    \vspace{-7mm}
    \end{subequations}
    \label{prob:follower trajectory problem}
    \end{problem}
    
    In words, the follower aims to minimize a cost objective \eqref{eq:follower:objective} that depends only on its own states and control, subject to dynamics and initial state constraints \eqref{eq:follower:dynamics}, state and static obstacle constraints \eqref{eq:follower:state constraint}, (state-dependent) control constraints \eqref{eq:follower:control constraints}, collision avoidance constraints \eqref{eq:follower:collision avoidance} and \eqref{eq:follower:collision avoidance final}, an inconvenience budget constraint \eqref{eq:follower:inconvenience budget}, and slack variable constraints \eqref{eq:follower:slack}.
    We describe three key elements of Problem~\ref{prob:follower trajectory problem} (highlighted in color) that are designed to encourage legible and proactive behaviors.
    
    \noindent\textbf{{\color{WildStrawberry}Markup term}}: Similar to a discount term common to many infinite horizon planning problems, in the cost function \eqref{eq:follower:objective}, we instead introduce the inverse of a discount factor: a \textit{markup} term $\mu> 1$ which increases the cost of states and controls further in the future, thereby incentivizing the follower to take nontrivial controls earlier in the (finite) horizon rather than later. For instance, the agent will be incentivized to start turning earlier to avoid the oncoming agent, therefore revealing its intent to pass on one side, rather than at the last possible moment. The hypothesis is that a markup $\mu>1$ will lead to more \textit{legible} and \textit{proactive} behaviors since the robot's intention to avoid other agents/obstacles will be revealed earlier in the interaction.

    \noindent\textbf{{\color{OliveGreen}Inconvenience budget}}: We constrain the amount of ``inconvenience'' an agent can experience via \eqref{eq:follower:inconvenience budget}, therefore ensuring no agent sacrifices too much on its own performance to benefit of others (hence preventing the onset of the frozen robot problem \cite{TrautmanKrause2010}). 
    We described the inconvenience budget in more detail in Section~\ref{subsec:inconvenience definition}.
    
    \noindent\textbf{{\color{RoyalBlue}Collision avoidance slack}}: To ensure Problem~\ref{prob:follower trajectory problem} remains feasible, we introduce a slack variable on the collision avoidance constraint \eqref{eq:follower:collision avoidance} with a weighting of $\gamma_0$. 
    Additionally, since the robot's model of the human within the IBR iterations will not match exactly with how humans truly behave, we apply a discount $\gamma < 1$ on the collision avoidance constraint so that collision constraint violations later in the planning horizon are penalized less. 
    Despite treating collision avoidance as a soft constraint, our experiments show that by the robot behaving legibly and proactively, there is better coordination between the human and robot which leads to more fluent and less collision-prone interactions.
    Additional safety filters can be added to further improve safety albeit in a reactive manner \cite{LeungSchmerlingEtAl2020,FisacAkametaluEtAl2018}, but this is outside the scope of this work.

    \subsection{Inconvenience budget}
    \label{subsec:inconvenience definition}
    
    Similar to \cite{SunZhanEtAl2018}, inconvenience is a quantity that measures the increase in cost of a planned trajectory compared to an ideal, or optimistic, trajectory. In this work, we consider the inconvenience experienced by the robot whereas \cite{SunZhanEtAl2018} considers the human inconvenience caused by the robot's actions.
    For notational simplicity, let $\traj{0:T+1}=(\xt{0:T+1},\ut{0:T})$ denote a trajectory of an agent (either human or robot). Then let $c(\traj{0:T+1})$ be a measure of how ``convenient'' a trajectory $\traj{0:T+1}$ is for some user-defined notion of convenience (lower values indicate more convenient trajectories). For example, convenience can be measured as a weighted linear combination of functions that includes trajectory length, jerk, and distance to the goal at the end of the trajectory, i.e., $c(\traj{0:T+1}) = w^T\phi(\traj{0:T+1})$.
    Note that $c$ does not necessarily need to be the same as the cost function $J$ in \eqref{eq:follower:objective}.
    Then we define the \textit{in}convenience of a trajectory $\traj{0:T+1}$ as,
    \begin{equation}
        J_\mathrm{incon}(\traj{0:T+1}) = \frac{c(\traj{0:T+1}) - c(\traj{0:T+1}_\mathrm{ideal})}{c(\traj{0:T+1}_\mathrm{ideal})},
        \label{eq:inconvenience}
    \end{equation}
    \noindent which describes the degradation in convenience of $\traj{0:T+1}$ compared to $\traj{0:T+1}_\mathrm{ideal}$ which is the solution to the \textit{idealized} trajectory optimization problem described in Problem~\ref{prob:ideal trajectory problem},
    
    \begin{problem}[Idealized trajectory optimization problem]
        \begin{subequations}
        {\small
    \begin{align}
    \min_{\xt{0:T+1},\ut{0:T}} & \sum_{t=0}^T J(\ut{t}, \xt{t}, t) + J_{T+1}(\xt{T+1}) \label{eq:ideal:objective}\\
        \mathrm{s.t.} \quad & \xft{t+1} = \ff(\xft{t}, \uft{t}), \quad  t=0,...,T \label{eq:ideal:dynamics}\\
        & \xft{0} = \xft{\mathrm{current}} \label{eq:ideal:initial state}\\
        & \xft{t} \in \mathcal{X}_\mathrm{F}^t \setminus \mathcal{O}_\mathrm{static}, \quad t=0,...,T+1, \label{eq:ideal:state constraint}\\
        & \uft{t} \in \mathcal{U}_\mathrm{F}(\xft{t}) , \quad t=0,...,T.   \label{eq:ideal:control constraints}
    \end{align}
    \label{eq:ideal formulation}
    }
    \vspace{-7mm}
    \end{subequations}
    \label{prob:ideal trajectory problem}
    \end{problem}
    
    \noindent Problem~\ref{prob:ideal trajectory problem} describes optimal trajectory towards a goal assuming no other agents are present (i.e., collision avoidance and inconvenience budget constraints are removed).
    
    \noindent \textit{Remark}: In the scope of Problem \ref{prob:follower trajectory problem}, $c(\traj{0:T+1}_\mathrm{ideal})$ is a \textit{constant} since it can computed independently of Problem~\ref{prob:follower trajectory problem}. Thus if $\phi$ consists of only convex functions (e.g., quadratic functions), then \eqref{eq:follower:inconvenience budget} is a convex (inequality) constraint.
    
    \subsection{Practical considerations when performing IBR}
    We highlight some important practical considerations.
    
    \noindent \textbf{Sequential Convex Program.} Generally, Problem~\ref{prob:follower trajectory problem} is nonlinear (due to dynamics and collision avoidance constraints). We can apply sequential convex programming (SCP), the process of repeatedly convexifying an optimization problem about a previous solution. We can linearize the dynamics and collision avoidance constraint, and if the convenience function (see Section~\ref{subsec:inconvenience definition}) and the cost function \eqref{eq:follower:objective} are quadratic, then Problem~\ref{prob:follower trajectory problem} is a quadratically constrained quadratic program, i.e., convex.
    
    \noindent \textbf{IBR initialization.} Performing IBR requires an initial trajectory for the leader at the first iteration. Instead of using the previous trajectory, we instead use the ideal trajectory for the current time step. This ensures the IBR solution at each time step will exhibit minimal deviations from the (current) ideal trajectory to the extent allowed by the inconvenience budget constraint. Initializing with the previous trajectory accumulates an imbalance between the human and robot (i.e., one agent would disproportionately move aside more).
    
    \noindent \textbf{Trust region.} Since we are using SCP where the linearization is valid locally around the point of linearization, we apply a trust region cost $J_\mathrm{trust}(\uf, \xf, \uf^\mathrm{prev}, \xf^\mathrm{prev}) = \beta (\| \uf - \uf^\mathrm{prev}\|_2^2 + \| \xf - \xf^\mathrm{prev}\|_2^2)  $ to \eqref{eq:follower:objective} to ensure the new solution does not differ too much from the previous solution. As such, combined with the IBR iterations, \textit{both} agents' trajectories will gradually deviate from their ideal trajectories until they avoid collisions or reach the inconvenience budget constraint.

    \subsection{Scaling up to multiple agents and wall constraints}
    Thus far, we have been concerned with a two-agent setting---one robot, one human.
    Extending IBR problems to account for all agents in the scene is possible but can be computationally challenging and typically relies on parallelization \cite{WangWangEtAl2021} to ensure real-time capabilities. 
    To reduce the number of agents to consider within the IBR loop, we lean on the idea that different agents have varying degrees of interactivity with the robot. For example, agents that are close to the robot but moving \textit{away} can be considered ``peripheral'' agents, whereas agents that are farther but moving towards the robot are considered ``interacting'' agents. 
    We consider the problem of defining an interactivity score or classifying which agents are interacting or ``peripheral'' outside the scope of this work but there are several works that investigate this problem \cite{KitazawaFujiyama2010,TopanLeungEtAl2022,TolstayaMahjourianEtAl2021}.
    Assuming interacting and peripheral agents have been classified, we propose a simple tractable way to account for both interacting and peripheral agents simultaneously---we treat peripheral agents as constant velocity obstacles which amount to additional \textit{linear constraints} (after linearization) to Problem~\ref{prob:follower trajectory problem}.
    Additionally, straight wall constraints (e.g., narrow corridor passing) are linear state constraints that can be easily appended to Problem~\ref{prob:follower trajectory problem}.

    \section{Experimental results and discussion}
    \label{sec:experiment and discussion}
    \subsection{Experimental set-up}
    \begin{figure*}[t]
        \centering
        \includegraphics[width=\textwidth]{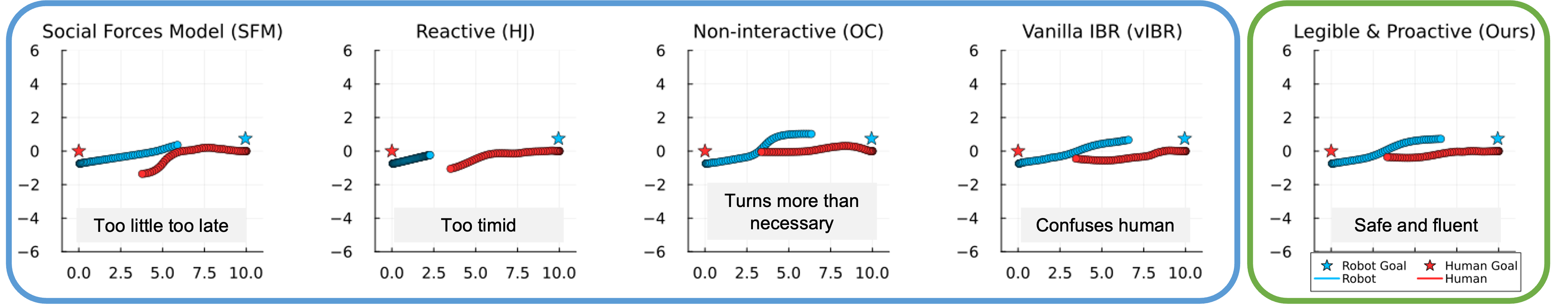}
        \caption{Human and robot trajectories in a head-on scenario. In our approach (green box) the robot legibly and proactively conveys its intent to pass, thus providing the human sufficient time to prepare to pass safely. In other approaches (blue boxes), the robot (SFM) does not convey its intention clearly and early and causes the human to swerve significantly, (HJ) freezes on the spot, (OC) takes on more collision avoidance responsibility than necessary, or (vIBR) confuses the human.}
        \label{fig:qualitative}
    \end{figure*}
    
    \begin{figure}
        \centering
        \includegraphics[width=0.48\textwidth]{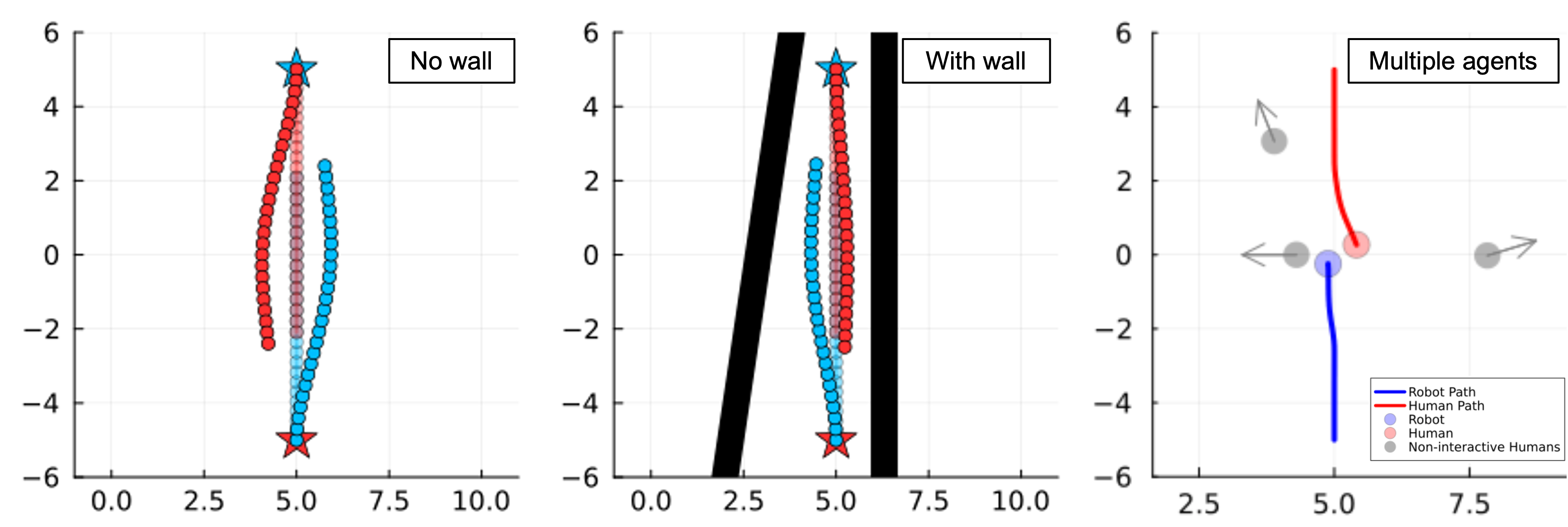}
        \caption{Our approach can easily account for additional wall constraints and multiple dynamic agents.}
        \label{fig:wall}
        \vspace{2mm}
    \end{figure}
    
    \begin{figure}[t]
        \centering
        \includegraphics[width=0.4\textwidth]{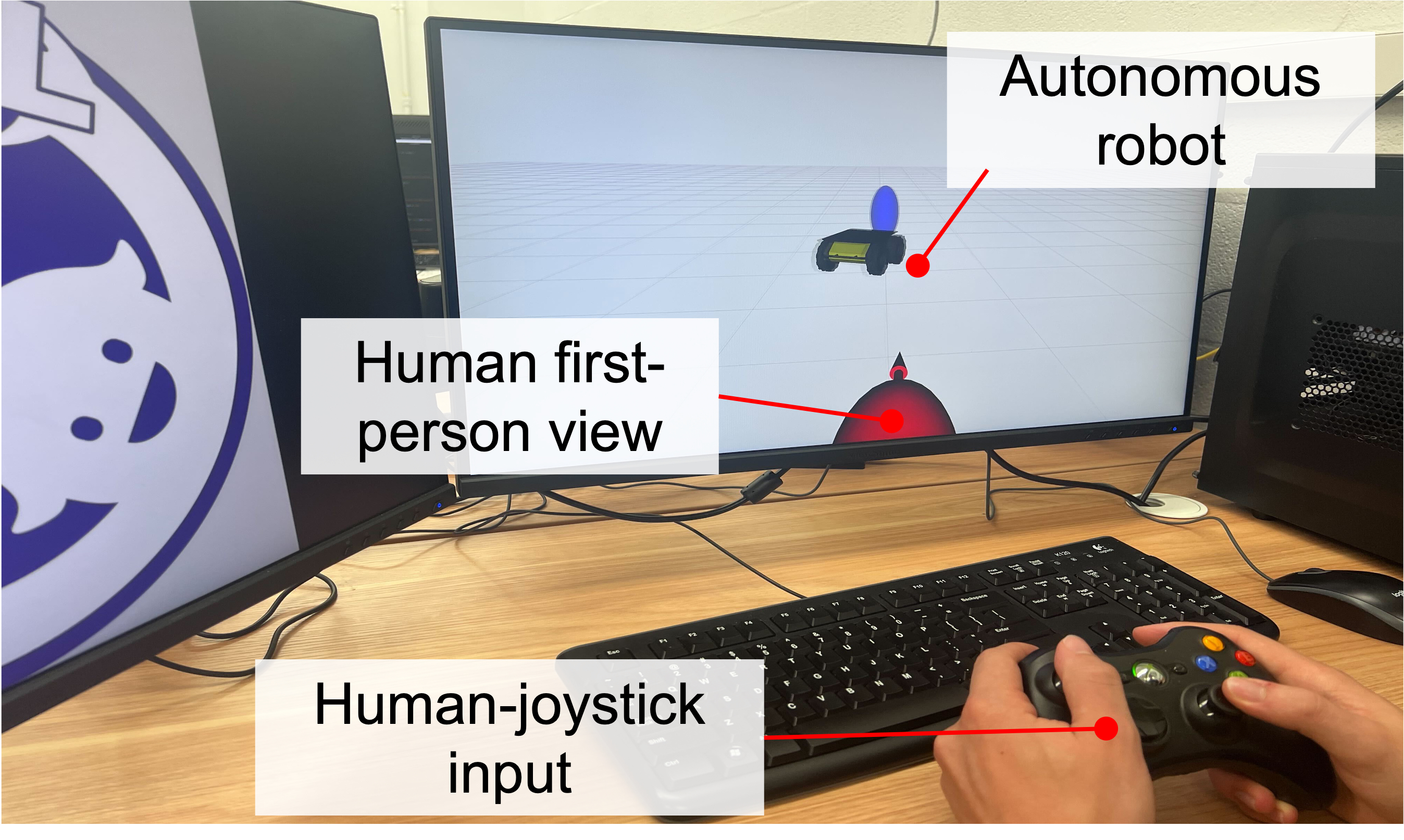}
        \caption{Snapshot of human-in-the-loop simulation experiment. Human interacts with the robot in first-person view.}
        \label{fig:hitl}
        \vspace{-3mm}
    \end{figure}
    
    We evaluate our proposed approach on a number of simulations and compare it against a number of baseline methods. We also perform human-in-the-loop experiments to demonstrate the real-time applicability of our method.
    
    \noindent \textbf{Software implementation.}
    We used the Julia programming language \cite{BezansonEdelmanEtAl2017}, and the JuMP package, a modeling language for a wide range of optimization problems \cite{DunningHuchetteEtAl2017,LubinDowsonEtAl2023}, to model the problem, and ECOS \cite{DomahidiChuEtAl2013} as the solver. We used forward-mode automatic differentiation \cite{RevelsLubinEtAl2016} to linearize the problem. For human-in-the-loop experiments, we used ROS \cite{ROS} to incorporate the human control input in real time. All experiments (except human-in-the-loop) were performed on an AMD Ryzen 9 3900X 12-Core Processor.
    
    \noindent \textbf{Dynamics.} Our method is dynamics-agnostic. For simplicity and ease of comparison, we use the dynamically-extended unicycle model for both the human and robot agents. The state and control (for each agent) are $\mathbf{x}~=~[x, y, \theta, v]^T$ and $\mathbf{u}~=~[\omega, a]^T$, and the continuous time dynamics are $\dot{\mathbf{x}}~=~[v\cos\theta, v\sin\theta, \omega, a]^T$. 
    We use the discrete-time dynamics (omitted for brevity) by applying zero-order hold on the continuous dynamics with a time step size of $\Delta t = 0.1$ seconds.
    The control and velocity limits are $\omega \in [-1,1]$rad and $a\in[-1.5, 1.5]$ms$^{-2}$, and $v\in[0,1.5]$ms$^{-1}$.
    
    \noindent \textbf{Comparison methods.}
    We compare our approach to several baseline methods.
    \textit{Social Forces Model (SFM)} \cite{HelbingMolnar1995}: A common multi-agent control method based on attractive and repulsive forces for goal reaching and obstacle avoidance. 
    \textit{Reactive Control via HJ reachability (HJ)}: A reactive control method that executes optimal collision avoidance control if a collision with the human is imminent under a constant velocity assumption \cite{VandenBergLinEtAl2008}. Hamilton-Jacobi reachability \cite{MitchellBayenEtAl2005,HJJAX2021} is used for collision checking and control.
    \textit{Optimal Control (OC)}: Problem~\ref{prob:follower trajectory problem} without IBR iterations, without markup ($\mu=1$), without the inconvenient budget constraint, and a collision slack weighting $\gamma_0$ of 1000.
    \textit{Vanilla IBR (vIBR)}: The OC method with three IBR iterations.
    \textit{Legible \& Proactive IBR (Ours)}: Problem~\ref{prob:follower trajectory problem} with 3 IBR iterations. %
        
    \noindent\textbf{Human simulation model.} To simulate the human in our experiment, we use two different approaches, IBR and OC (as described above). We assume the human's goal is known, which is privileged information. We also add some white noise to the human controls. The parameters used to model the human is different from the robot's model of the human to make sure the robot does not use privileged information. 
    
    \noindent \textbf{Experimental parameters.}
    We considered a planning horizon of 2.5 seconds, and each episode was 5 seconds. When comparing against baselines, we used an inconvenience budget of 0.2, markup $\mu=1.05$, collision discount $\gamma=0.98$, a collision slack weighting $\gamma_0=150$, and a collision radius of 1 meter. We use a quadratic cost function, penalizing only running control cost and distance to the goal position. 
    For the inconvenience budget, we considered a linear combination of the sum of distances between states along the trajectory squared, the sum of the change in velocity squared, and the distance to the goal squared. Note that these quantities are quadratic.
    The human and robot approach each other almost head-on. Both agents have their goal position situated 10 meters directly in front of them, but their relative heading varies between $\pm \frac{\pi}{4}$.
    Their straight-line trajectory between the initial and goal state intersects, meaning they must deviate from the straight-line path to avoid collision.
    
    \subsection{Qualitative analysis}
    
    \begin{figure*}[t!]
        \centering
        \begin{subfigure}[t]{0.65\textwidth}
            \centering
            \includegraphics[width=0.48\textwidth]{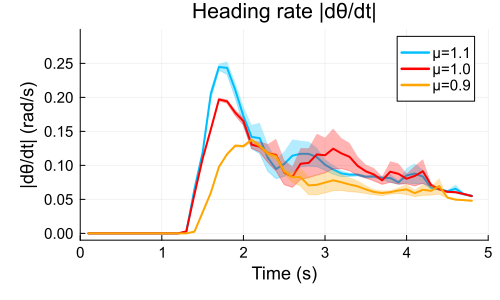}
            \includegraphics[width=0.48\textwidth]{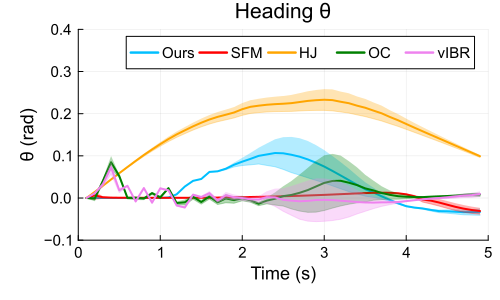}
            \caption{Left: Varying the markup parameter affects how quickly the robot starts turning to pass an oncoming human. Right: With our method, the robot's heading is consistent, smooth, and deliberate, whereas other methods oscillate between passing left or right or swerving too much.}
            \label{fig:proactive}
        \end{subfigure}%
        ~ 
        \begin{subfigure}[t]{0.33\textwidth}
            \centering
            \includegraphics[width=\textwidth]{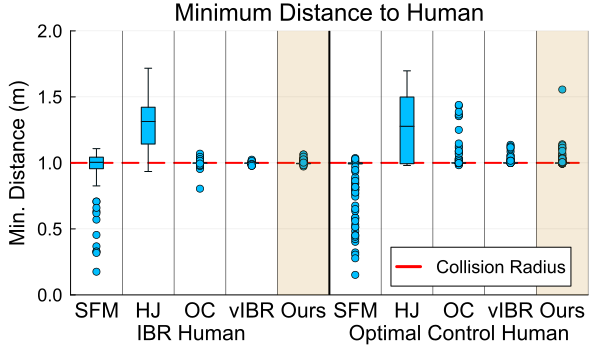}
            \caption{Box plots comparing the minimum distance between agents. Our approach does exceed the collision radius threshold despite using less stringent safety measures compared to other approaches.}
            \label{fig:safety}
        \end{subfigure}
        \vspace{5mm}
        \caption{(a) Analysis of heading and heading rate to illustrate the legibility and proactivity of our proposed approach. (b) Statistics of the minimum distance between human and robot to evaluate the safety performance of our approach.}
    \end{figure*}

    \noindent \textbf{General behavior.} Figure~\ref{fig:qualitative} illustrates a canonical scenario where the robot and human approach each other head-on---if the agents do nothing, collision is inevitable. We see that in our method (right, green box), both agents engage cooperatively in joint collision avoidance. 
    The reason is that the robot has indicated early and clearly its intent to pass on the right. 
    Whereas in SFM, the robot does not move out of the way until the last moment and does so very slowly, causing the human to make a sharp swerve at the last moment. 
    In the HJ approach, the constant velocity assumption is too conservative and causes the robot to almost immediately freeze on the spot, resulting in the human doing most of the collision avoidance. 
    In the OC case, does not anticipate the human will cooperate and makes a sharp turn to avoid a collision.
    In the vIBR case, while both agents deviated to avoid collision, the robot's motion was not legible or proactive which caused the human to sway back and forth a bit before passing on the left.
    
    \noindent \textbf{Additional constraints.} We tested our approach with wall constraints and with multiple agents. Figure~\ref{fig:wall} (left) compares the trajectories with and without the presence of a wall, while the figure on the right illustrates our planner navigating through a crowd. The human is forced to swerve more than the robot since the robot is blocked by another (non-interacting) agent. When adding 4 additional non-interacting agents, the computation time increased by $\sim6\%$, from approximately 67ms to 71ms per planning time step.
    
    \noindent \textbf{Effect of markup.} Figure~\ref{fig:proactive} illustrates how the robot is proactive and legible through its heading and heading rate. In the left plot, as markup increases, the robot rotates faster, thus making the robot's decision to move to one side more noticeable earlier. In the right plot, with our proposed method, the robot's heading is smoother, more deliberate, and less aggressive compared to HJ. Whereas SFM barely turns until the end, and OC and vIBR are indecisive and only start turning later in the interaction.
    
    \noindent \textbf{Human-in-the-loop simulation.}
    We performed human-in-the-loop experiments, see Figure~\ref{fig:hitl}, to validate the performance of our planner and demonstrate its real-time applicability. The planning was performed at 10Hz on a laptop with AMD Ryzen 7 PRO 6850U processor.

    \begin{figure*}[h]
        \centering
        \includegraphics[width=0.32\textwidth]{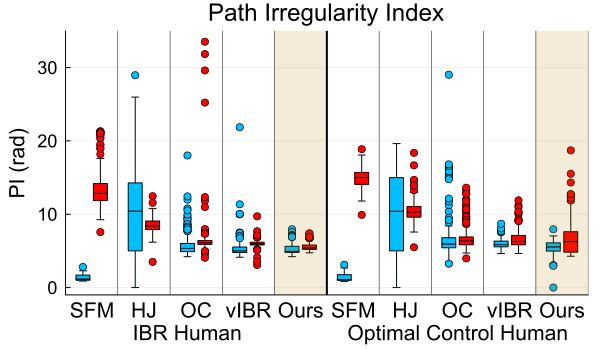}
        \includegraphics[width=0.32\textwidth]{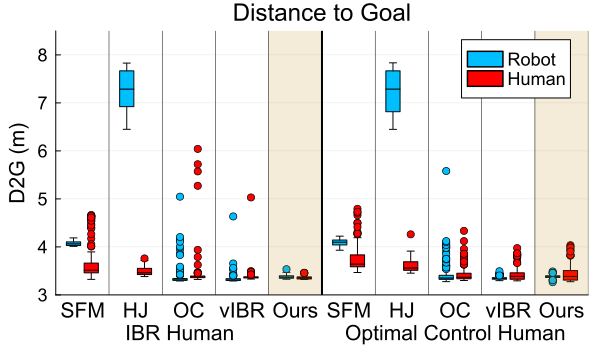}
        \includegraphics[width=0.32\textwidth]{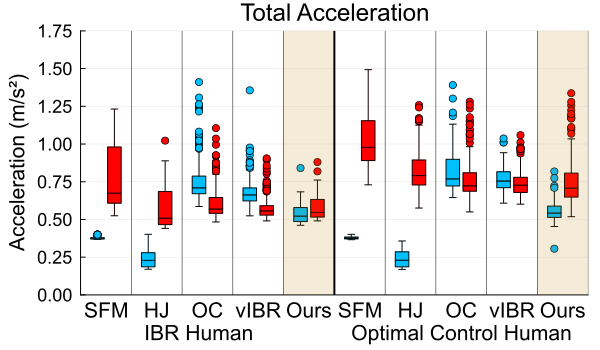}
        \caption{Box plots comparing efficiency performance metrics using two different human models (interactive and non-interactive). Our approach has a lower and more equitably distributed between the human and robot. (Lower is better.)}
        \label{fig:efficiency metrics}
    \end{figure*}

    \subsection{Performance metrics}
    \label{subsec:metrics}
    We consider the following metrics to evaluate the safety and efficiency of our approach.
    
    \noindent \textit{Minimum distance (MinDist)}: To evaluate safety, we consider the minimum distance between the robot and human during the interaction. Let $\mathrm{pos}(\cdot)$ denote the position vector given agent state. Then the minimum distance is defined as,
    
    \vspace{-1mm}
    {\small
    \begin{equation*}
        \mathrm{MinDist}(\xht{0:T+1}, \xrt{0:T+1}) = \min_{t\in\{0,...,T+1\} } \| \mathrm{pos}(\xht{t}) - \mathrm{pos}(\xrt{t})\|_2
    \end{equation*}
    }
    \noindent \textit{Path Irregularity (PI) Index} \cite{GuzziGiustiEtAl2013}: The total amount of unnecessary turning relative to the straight line path towards the goal. Lower is better. Let $\mathbf{v}_t=[v_t\cos\theta_t, v_t\sin\theta_t]^T$ be the velocity vector at time $t$, and $\mathbf{d}_t^\mathrm{str}=[x_\mathrm{goal} -x_t, y_\mathrm{goal}-y_t]^T$ be the vector pointing to the goal from the position at time $t$. Then the path irregularity index is defined as,
    \begin{equation*}
        \mathrm{PI}(\tau^{0:T+1}) = \sum_{t=0}^{T+1} \cos^{-1}\left(\frac{\mathbf{v}_t \cdot \mathbf{d}_t^\mathrm{str}}{\|\mathbf{v}_t \| \| \mathbf{d}_t^\mathrm{str}\|}\right)
    \end{equation*}
    
    \noindent \textit{Distance to goal (D2G)}: To evaluate how well an agent accomplishes its task, we compare how close the agent gets to the desired goal location by the end of the horizon. Lower is better. Note: All simulations have the same horizon length. 
    \begin{equation*}
        \mathrm{D2G}(\tau^{0:T+1}) =  \| \mathrm{pos}(\xt{T+1}) - \mathrm{pos}(\xt{\mathrm{goal}})\|_2
    \end{equation*}
    \noindent \textit{Total acceleration (ACC)}: To evaluate the amount of effort and smoothness of the trajectory, we compute the total acceleration over the trajectory. Lower is better.
    \begin{equation*}
        \mathrm{ACC}(\tau^{0:T+1}) = \frac{1}{\Delta t} \sum_{t=0}^{T} \| \mathbf{v}_{t+1} - \mathbf{v}_t\|_2
    \end{equation*}    
    \vspace{-3mm}
    
    \subsection{Quantitative analysis}
    We compare our approach against the baseline methods using the performance metrics described in Section~\ref{subsec:metrics}.
    
    \noindent \textbf{Safety.} Figure~\ref{fig:safety} compares the minimum distance between the robot and human over the interaction episode. We see that our method does not violate the collision radius threshold despite collision avoidance not being a hard constraint, and having a lower slack penalty than OC. In the OC case where the weighting on the slack variable was much higher than Ours, there were a few cases where the minimum distance dipped below the collision radius. This result is consistent with our hypothesis---if the robot is more proactive and legible, then the human can prepare, and both can engage equitably in avoiding collision.
    
    \noindent \textbf{Efficiency.} Figure~\ref{fig:efficiency metrics} compares the efficiency-related performance metrics. 
    Both SFM and HJ methods result in a large imbalance in all efficiency metrics, often requiring the human to bear a majority of the collision avoidance responsibility, i.e., swerving a lot. This is because SFM moves aside too late which is ineffective in conveying intent to the human, while HJ is too timid and often freezes on the spot requiring the human to swerve out of the way.
    OC, in general, performs similarly to vIBR and Ours, but experiences large fluctuations since it does not account for the interaction dynamics---it does not anticipate how the human may respond, and therefore often gets in the way of the human, requiring one or both to swerve out of the way (see Figure~\ref{fig:qualitative}).
    vIBR performs similarly to Ours, but Ours either has a lower variance, is more equitable, or for roughly the same human performance, the robot does better (especially notable in ACC).
    These results indicate that with relatively simple modifications to the standard trajectory optimization problem, we are able to achieve more prosocial (i.e., equitable) and more efficient robot behaviors without compromising on safety.

    \section{Conclusions and future work}
    \label{sec:conclusion}
    We have presented simple yet effective modifications to a general trajectory optimization problem to generate legible and proactive robot behaviors for prosocial human-robot interactions. With our framework, we are able to achieve more efficient and more equitable interactions without compromising on safety. Our human-in-the-loop experiments highlight the real-time applicability of our approach. We also propose a simple and scalable approach to account for multiple agents but requires future work to distinguish which humans are considered interacting or peripheral.
    Future work is aimed at scaling our efforts to account for more humans and perform real-world human-in-the-loop testing. To ensure scalability, we plan to explore different interaction-aware planning frameworks, such as using neural-based human behavior prediction models to model interaction.
\bibliographystyle{IEEEtran}
\bibliography{../../../bib/main,../../../bib/ctrl_papers}

\end{document}